\documentclass[letterpaper, 10 pt, conference]{ieeeconf}  

\IEEEoverridecommandlockouts                              

\usepackage{graphics} 
\usepackage{epsfig} 
\usepackage{mathptmx} 
\usepackage{times} 
\usepackage{amsmath} 
\usepackage{amssymb}  
\usepackage{subfigure}
\usepackage{algorithmic}
\usepackage{algorithm}
\usepackage{bbm} 
\usepackage{color}
\usepackage[obeyspaces]{url}
\usepackage{multirow}
\DeclareMathOperator*{\argmin}{arg\,min} 

\title{\LARGE \bf
On Support Relations and Semantic Scene Graphs
}

\author{Wentong Liao$^{1}$, Michael Ying Yang$^{2}$, Hanno Ackermann$^{1}$ and Bodo Rosenhahn$^{1}$
\thanks{$^{1}$Wentong Liao, Hanno Ackermann and Bodo Rosenhahn are with Institute of Information Processing
        Leibniz University Hannover, Germany}%
\thanks{$^{2}$Michael Ying Yang (Corresponding Author) is with University of Twente-ITC, Netherlands
        {\tt\small (michael.yang@utwente.nl)}}%
}

\begin{document}

\maketitle
\thispagestyle{empty}
\pagestyle{empty}

\begin{abstract}
Rapid development of robots and autonomous vehicles requires semantic information about the surrounding scene to decide upon the correct action or to be able to complete particular tasks. Scene understanding provides the necessary semantic interpretation by semantic scene graphs. For this task, so-called support relationships which describe the contextual relations between parts of the scene such as floor, wall, table, etc, need be known. This paper presents a novel approach to infer such relations and then to construct the scene graph. Support relations are estimated by considering important, previously ignored information: the physical stability and the prior support knowledge between object classes. In contrast to previous methods for extracting support relations, the proposed approach generates more accurate results, and does not require a pixel-wise semantic labeling of the scene. The semantic scene graph which describes all the contextual relations within the scene is constructed using this information. To evaluate the accuracy of these  graphs, multiple  different measures are formulated. The proposed algorithms are evaluated using the NYUv2 database. The results demonstrate that the inferred support relations are more precise than state-of-the-art. The scene graphs are compared against ground truth graphs.
\end{abstract}

\section{INTRODUCTION}
Scene understanding is a popular but challenging topic in computer vision, robots and artificial intelligence. It can be roughly divided into object recognition~\cite{uijlings2013selective}, 
layout estimation~\cite{choi2013understanding}, and physical relations inference~\cite{Mottaghi2015arXiv}. 
Traditional scene understanding mainly focuses on object recognition and has achieved great developments, especially by recent developments in deep learning~\cite{krizhevsky2012imagenet}.
Exploring more vision cues like contextual and physical relations between objects is becoming the topic of great interest in the computer vision community. In many robotic applications as well, knowledge about relations between objects are necessary for a robot to finish its task. For example, for a robot to take a newspaper from under a cup, it must first lift the cup, and then put it back or place it somewhere else.


A semantic scene graph is an effective tool for representing physical and contextual relations between objects and scenes. In \cite{wu2014hierarchical} it was proposed to use a semantic scene graph in a robotic application. 
Scene graphs have also been used in different  applications~\cite{koppula2011semantic,johnson2015image} and \cite{schuster2015generating}. However, in most existing works, a scene graph is regarded as input for scene understanding.

Our goal in this work is to infer reasonable support relations and then to generate a  semantic scene graph. 
To accurately estimate support relations, we propose a framework based on object detection and contextual semantics instead of pixelwise segmentation or 3D cuboids which are used in previous methods for inferring support relations. 

\begin{figure*}[thpb]
      \centering
      \framebox{\includegraphics[width=0.87\textwidth]{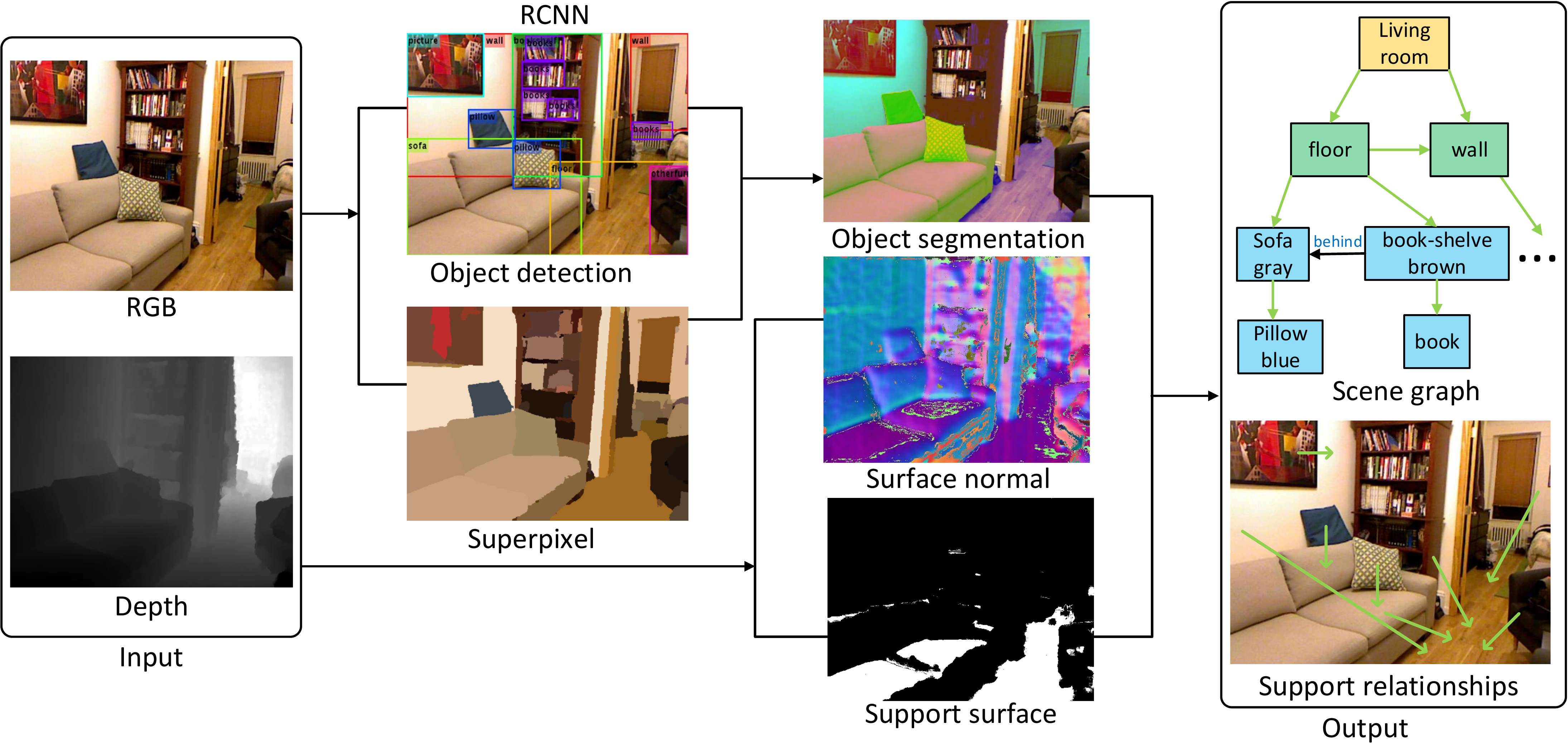}}
      \caption{\textbf{Overview.} Given an RGBD image, we use a CNN for object detection. A superpixel map is also computed. \textit{Bounding boxes} of detected objects and the \textit{superpixel map} are used to segment objects. Parallel to this process, surface normals, support surface and 3D point cloud aligned to the room are computed. Physical support between objects are estimated and a scene graph is inferred. Green arrows indicate the relation from the supported object to the surface that supports it.}
      \label{fig:pipeline}
   \end{figure*}

With the information achieved from scene recognition, object recognition, attribute recognition, support estimation and relative spacial estimation, a semantic graph is inferred to describe the given scene. Additionally, we introduce some metrics to evaluate the quality of a generated semantic graph, an issue so far not considered. An overview of our approach is illustrated in Fig.~\ref{fig:pipeline}. 


We analyze our method on the benchmark dataset NYUv2 of cluttered room scenes. 
The results show that our algorithm outperforms the state-of-the-art for support inference. 
Quantitative and qualitative comparisons with ground truth scene graphs show that the estimated graphs are accurate.

To summarize, our contributions are:
\begin{itemize}
\item We propose a new method for inferring more accurate support relations compared to previous works \cite{silberman2012indoor,xue2015towards}.
\item Neither pixel-wise semantic labelings nor 3D cuboids are necessary.
\item We introduce a way how to construct semantic scene graphs and assess the quality
\item Ground truth scene graphs of the NYUv2 dataset are provided to the scientific community
\item A convenient GUI tool for generating ground truth graphs will be made  available\footnote{More information on the tool can be found in the supplementary material of this paper. This information will be provided on the authors' homepage.}
\end{itemize}

This paper is structured as follows: related work is discussed in Sec.~\ref{Sec:Related.Work}. Object recognition and segmentation, features for scene and support relations classification  are shortly explained in Sec.~\ref{Sec:Detection.Classification}. In 
Sec.~\ref{Sec:Support}, the model for support inference is proposed. 
How a scene graph is inferred can be explained in Sec.~\ref{Sec:Scene.Graph}. Experimental results of the proposed 
framework are shown in Sec.~\ref{Sec:Exps}. Finally, a conclusion in 
Sec.~\ref{Sec:Conclusion} summarizes this paper.


\section{Related Work}
\label{Sec:Related.Work}
Justin et al.~\cite{johnson2015image} proposed to use scene graphs as queries to retrieve semantically related images. Their scene graphs are manually generated by the Amazon Mechanical Turk, which literally is expensive.
Prabhu and Venkatesh~\cite{prabhu2015attribute} constructed scene graphs to represent the semantic characteristics of an image, and used it for image ranking by graph matching. Their approach works on high-quality images with few objects. 
Lin et al.~\cite{lin2014visual} proposed to use scene graphs for video search. 
Their semantic graphs are generated from text queries using manually-defined rules to transform parse trees, similar as~\cite{schuster2015generating}. 
Using a grammar, Liu et al.~\cite{liu2014creating} proposed to learn scene graphs from ground truth graphs of synthetic data. Then they parsed through a pre-defined segmentation of a synthetic scene so as to create a graph that matches the learned structure. 
None of these works objectively assess the quality of scene graph hypotheses compared with ground truth graphs. However, reasonable measures for this problem are important
especially after the publication of the \emph{Visual Genome} dataset~\cite{krishna2016visual}.

Physical relations between objects to help image or scene understanding have been investigated in~\cite{silberman2012indoor,xue2015towards,jia20133d,zheng2015scene}, and \cite{wong2015smartannotator}. 
Pixel-wise segmentation and 3D volumetric estimation are two major methods for this task.
\cite{silberman2012indoor,xue2015towards} used pixel-wise segmentations to analyze support relations in challenging cluttered indoor scenes. They both ignored the contextual knowledge provided by the scene. Silberman et al. \cite{silberman2012indoor} ignored small objects and the physical constraints while Xue et al.~\cite{xue2015towards} set up simple physical constraints.
A typical examples of 3D cuboid based method is~\cite{jia20133d}.
Jia et al. estimated the 3D cuboids to capture spatial information of each object using RGBD data and then reason about their stability. However, stability and support relations are inferred in tiny images with few objects. 

For the part of support relations inference in this paper is mostly related to \cite{silberman2012indoor,xue2015towards}. However, we integrate physical constraints and prior support knowledge between object classes into our approach for extracting more accurate support relations. Furthermore, we do not operate pixelwise segmentation for object extraction.
Finally, our framework generates a semantic graph to interpret the given image. Objective measures for accessing the quality of constructed graphs are proposed.


\section{Object detection and classification}
\label{Sec:Detection.Classification}
   \begin{figure*}[thpb]
	\centering
	\framebox{
	\subfigure[Object detection]{
	\label{fig:two_proposal}
	\includegraphics[width=0.25\textwidth]{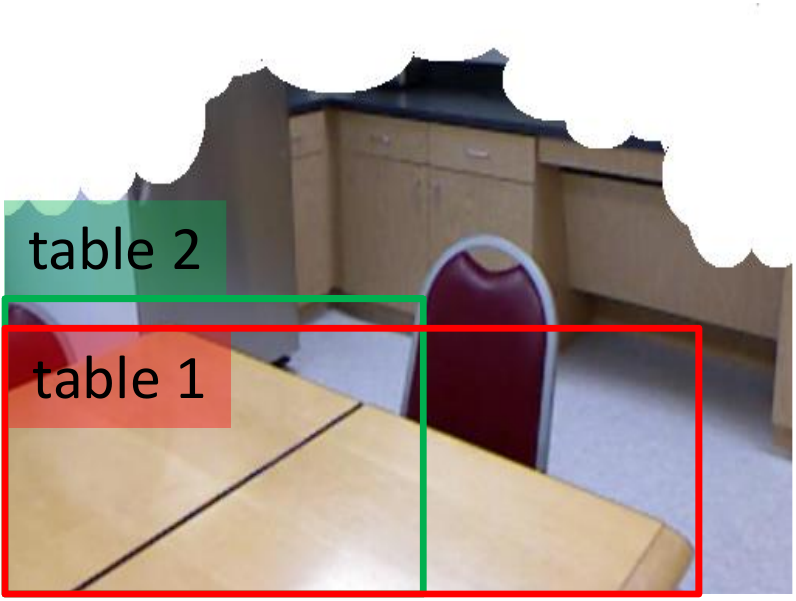}}
	\subfigure[Object segmentation]{
	\label{fig:object_seg}
	\includegraphics[width=0.55\textwidth]{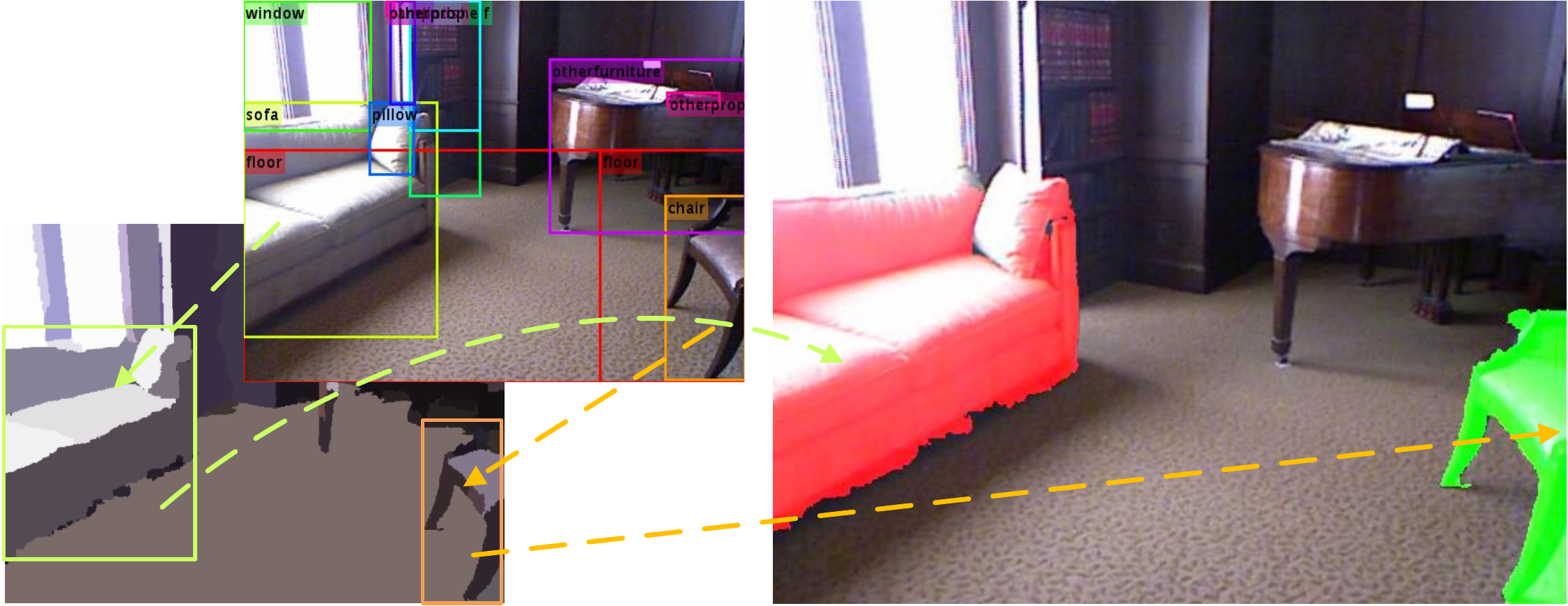}}
	}
	\caption{Examples of object detection (a) and object segmentation from bounding box and superpixel map (b). (a) two of the proposals with large intersection are classified as table. (b) The left two images show detected objects in bounding boxes from in RGB image (left-up) and corresponding superpixel map (left-down). The right image illustrates the results of object segmentation (only shows sofa and chair).}
\label{fig:proposal_selection}
   \end{figure*}

Recently, deep learning based algorithms have shown great success in object detection and classification tasks~\cite{deng2009imagenet,krizhevsky2012imagenet}, and \cite{gupta2014learning}. 
Here, the RCNN framework proposed by Gupta et al.~\cite{gupta2014learning} is applied to recognize objects in an image, which utilizes HHA, abbr. of horizontal displacement (depth), height and angle of the pixel’s local surface normal, representations to enhance object classification ability. 
In the RCNN, a pool of candidates is created and then each is indicated by a bounding box and a confidence score $sb$. 
Then, a class specific CNN assigns each proposal a classification score $sc$.
Finally, a non-maximum suppression is used to remove overlapping detections and obtain the final object proposals. 
Different from the original work, a weighted score $sw_i$ is used in the final step to obtain better bounding boxes of detected objects
\begin{equation}
 sw_i = sb_i + w*sc_i,  ~~i\in 1...N,
\end{equation}
where $w$ is the weight factor and $N$ is the total number of detected objects. 
Fig.~\ref{fig:two_proposal} shows an example of proposal decision. 
The proposals (in green and red bounding boxes respectively) are two out of several proposals that are classified as table and have large intersection. 
The red bounding box is decided using the weighted score in the non-maximum suppression step while the green box is the result of \cite{gupta2014learning}. It can be seen that the red one covers the table more accurately than the green one. This result is important in the following step of estimating relative positions between objects.


\subsection{Object Segmentation}
Correct separation of foreground objects from the background in the bounding box is critical for estimating accurate support relationships.
Many approaches can effectively complete this task, such as Grabcut~\cite{rother2004grabcut,lempitsky2009image}, semantic segmentation \cite{long2015fully} and instance segmentation \cite{gupta2014learning}.
But they are very computation time costly: Grabcut takes minutes for each bounding box, and deep learning-based approaches need several days for training and still several minutes for each input image. Furthermore, they require high performance GPUs which
is a limitation for practical applications, e.g. robots.  
We propose a simple and fast method to segment objects based on superpixel maps. 

Starting with the smallest (area) bounding box and continuing in ascending order, superpixels are assigned to bounding boxes if more than a certain ratio (in this paper: $80\%$) of their area is within the box.
Fig.~\ref{fig:object_seg} shows an example of segmentation result of our method: the sofa and chair are well separated from the background. This method is very efficient in terms of computation time and power. 

The superpixel maps used in this step are produced by the RCNN during the object recognition process. In other words, no additional computation is required to generate superpixel map for each image.


\subsection{Scene and support relations classification}
Indoor scene category is an important auxiliary information for object recognition, for instance, a bathtub is impossible in a living room, as shown in Fig. \ref{fig:prior_obj_scene}. In this paper, we use the spatial pyramid formulation of \cite{lazebnik2006beyond} as feature for this task and use logistic regression classifier to make probabilistic prediction. 

We furthermore benefit from using this method, which the SIFT descriptors generated in this process are used as the feature for classifying support relations, then computation is saved by extracting specific features. A logistic regression classifier $D_{SP}$ is trained with features $F^{SP}_{i,j}$ associating with support label $y^s\in\{1,2,3\}$ to indicate object $j$ supports object $i$ from below, behind, and no support relationship, respectively. Note that the features are asymmetric, i.e. $F^{SP}_{i,j}$ is not for judging if object $i$ supports object $j$.


\subsection{Coordinate Alignment}

A suitable coordinate system is necessary for correctly estimating object positions. 
Therefore, the image coordinates need to be aligned with the 3D room coordinates first. We find the principle directions $(\mathbf{v}_1,\mathbf{v}_2,\mathbf{v}_3)$ of the aligned coordinate based on the Manhattan world assumption \cite{coughlan2003manhattan} :most of the visible surfaces are located along one of three orthogonal directions. 
The "wall" or "floor/ground"  (We note the floor as ground for convenient interpretation in the follows of this paper)
detected by RCNN (if yes) indicates the horizontal or vertical direction of the scene. This useful cues are embodied to our method for coordinate alignment.
Each pixel has image coordinates $(u,v)$, 3D coordinates $(X,Y,Z)$, and the local surface normal $(N_x,N_y,N_z)$. 
As discussed in \cite{silberman2012indoor}, straight lines are extracted from images and the mean-shift modes of surface normals are computated. 
For each line that is very closed to $Y$ direction, two other orthogonal candidates are sampled for computing the score as follows: 
\begin{align}
&S(\mathbf{v}_1,\mathbf{v}_2,\mathbf{v}_3) = \sum^3_{j=1}(SN_j+SL_j)~~~j= 1,2,3\\
&SN_j=\frac{w_N}{N_N}\sum^{Num_N}_iexp(\frac{-(N_i*V_j)^2}{\sigma^2}+\mathbb{I}(y_{N_i})P_{N_i})\\
&SL_j=\frac{w_L}{N_L}\sum^{Num_L}_iexp(\frac{-(L_i*V_j)^2}{\sigma^2}).
\end{align}
Here, $\mathbf{v}_1,\mathbf{v}_2,\mathbf{v}_3 $ indicate the three principal directions,  $N_i$ the normal of a pixel, $L_i$ the direction of a straight line, $Num_N$ and $Num_L$ the number of points and lines on each surfacce, respectively, and $w_N$ and $w_L$ the  weights of the 3D normals and line scores, respectively. 
$\mathbb{I}(y_{N_i}) = 1$, if the region which includes the pixel ${N_i}$ is "ground" or "wall" and $P_{N_i}$ is the corresponding predicted probability, else $\mathbb{I}(y_{N_i}) = 0$. This term favors the candidate that is most perpendicular to the ground or wall surface to be chosen as one of the principle direction and further ensures the ground to be the lowest surface.
The candidates $(\mathbf{v}_1,\mathbf{v}_2,\mathbf{v}_3)$ which have the maximal score are chosen as the aligned coordinate system and the one of the three directions which is closest to the original $Y$ direction is chosen as $\mathbf{v}_y$. Then the image coordinate is aligned to $(\mathbf{v}_x,\mathbf{v}_y,\mathbf{v}_z)$.
 


\section{Modeling Support Relationships}
\label{Sec:Support}
Given an image with $N$ detected objects with class labels $\textbf{C}=\{C_1,\dots,C_N\}$, 
then (a) the visible supporter of object $C_i$ is denoted by $S_i\in\{i=1 \dots N\}$, (b) $S_i=N+1$ indicates that object $C_i$ is supported by an invisible object and (c) $S_i=ground$ means that $C_i$ is the ground and does not need support. 
The supporting type of $C_i$ is encoded as $ST_i = 1$ for being supported from behind and $ST_i = 0$ for the support from below.

The following assumptions are used in our model: Every object is either
(a) supported by another detected object next to it in the image plane, in which case $S_i\in\{1\dots N\}$,
(b) supported by an object not detected or invisible in the image plane, or $S_i=N+1$,
(c) it is ground itself which requires no support $S_i=ground$.

\begin{figure}[t!]
	\centering
	\framebox{\includegraphics[width = 0.46\textwidth]{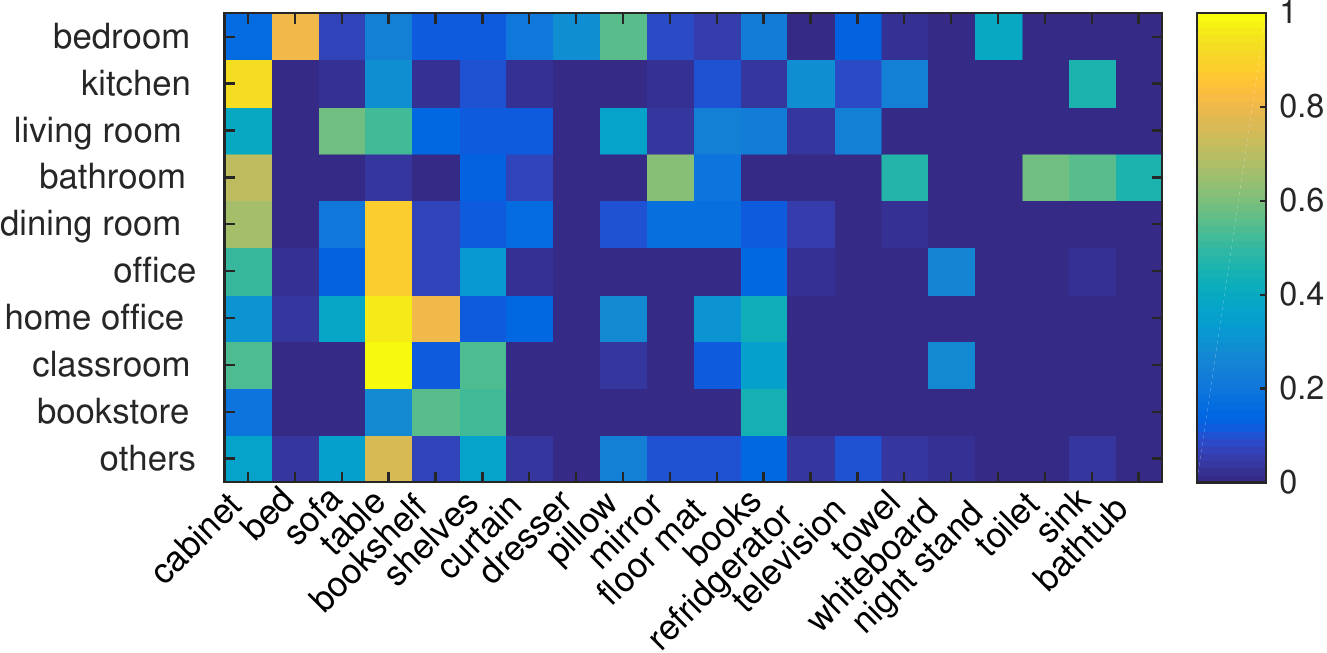}}
	\caption{Prior knowledge of specific class object presenting in specific scene type. Not all of the object categories are shown here.}
	\label{fig:prior_obj_scene}
\end{figure}

In practice, the object classes and physical factors (e.g. physical rationality and common sense) constrain the support relations of indoor scenes. For example, the window behind the sofa in Fig. \ref{fig:object_seg} is supported by the wall rather than other objects or parts of the scene. However, 
without such common sense rules, it is more likely to be supported by the sofa.
To infer more accurately support relationships, the object classes information and some constrains are added to our model.


We infer the support relationships similarly as in~\cite{silberman2012indoor}: the most probable joint assignment of support object $\textbf{S} = \{S_i \dots S_N\}$, support type $ST\in\{0,1\}$ and object classes $\textbf{C}=\{C_1,\dots,C_N\}$
\begin{equation}
	\{\textbf{S}^*,\textbf{ST}^*,\textbf{C}^*\} = \argmin_{\textbf{S},\textbf{T},\textbf{C}}E(\textbf{S},\textbf{T},\textbf{C}).
\label{eq:obj_func}
\end{equation}
The energy of our model in Eq. \eqref{eq:obj_func} is divided into four parts: the support energy $E_{SP}$, the object classification $E_C$, and the physical constraint energy $E_{PC}$.
The total energy function is formally defined as:
\begin{equation}
	E(S,ST,SC) = E_{SP}(S,ST) + E_C(C)+ E_{PC}(S,ST,C),
\label{eq:energy}
\end{equation}
where
\begin{align}
	E_{SP}(S,ST) &= -\sum^N_i log(D_{SP}(F^{SP}_{i,j}|S_i,ST_i)),\label{eq:energy.SP}
	\\
	E_C(C) &= -\sum^N_i log \{P_{C_i}P(C_i|SC)D_{SC}(F^{SC}|SC)\}.
\label{eq:energy.C}
\end{align}		
Here, $D_{SP}$ is the trained support classifier, $F^{SP}_{i,j}$ are the support features for $C_j$ supporting $C_i$, $P_{C_i}$ is the object category probability of $C_i$ predicted by the RCNN, $P_{(C_i|SC)}$ is the probability of object class $C_i$ being present in the scene. $D_{SC}$ is the trained scene classifier and $F^{SP}_{i,j}$ are the features.
Fig.~\ref{fig:prior_obj_scene} shows theprior knowledge of 20 object classes in the dataset. For instance, $P(bed|bedroom) = 0.9$ means that $90\%$ of images taken in bedroom do have a bed.

The physical constraint energy $E_{PC}$ consists of several items:
(1) The \textbf{Object class constraint $C_C$:} This object class constraint is imposed onto the support relations of a given object. For any support object, its lowest point should not be higher than the highest points of the supported object. The ground needs no support and must be the lowest points in the aligned coordinate system, 
\begin{equation}
C_C(S_i,C_i) = \left\{
\begin{aligned}
&-logP^{SP}_{C_{S_i},C_i},~if~C_i \neq ground~AND~H^b_{S_i}\leq H^t_i,\\
&-logP(C_i),~if~C_i = ground~AND~\\
&~~~~~~~~~~~~~~~~~~~~~~~H^b_j>H^b_i,~\forall C_j\\
&\infty,~otherwise.
\end{aligned}
\right.
\label{eq:obj_class_cons}
\end{equation}
Here, $H^b_i$ and $H^t_i$ are the lowest and highest points in aligned 3D coordinates of object $C_i$, respectively, and $P^{SP}_{C_{S_i},C_i}$ encodes the prior of object class $C_{S_i}$ supporting object class $C_i$ (but not vice versa). 

(2) The \textbf{Distance constraint $C_{dist}$:} for any object, its supporter must be adjacent to it to satisfy the principle of physical stability. Formally, the  distance constraint is defined as:
\begin{equation}
C_{dist}(S_i,C_i,ST_i) = \left\{
\begin{aligned}
&(H^b_i-H^t_{S_i})^2,~if~ST_i=1,\\
&V(S_i,C_i),~~~~if~ST_i=0,\\
\end{aligned}
\right.
\label{eq:dist_cons}
\end{equation}
where $V(S_i,C_i)$ is the minimum horizontal distance from $C_i$ to its supporter $S_i$.

(3) The \textbf{Support constraint $C_{SPC}$:} Besides the ground, all detected objects must be supported, and no object is lower than the ground. This constraint is formally defined as:
\begin{equation}
C_{SPC}(S_i,C_i) = \left\{
\begin{aligned}
&\infty,~if~S_i=ground~AND~C_i\neq ground\\
&\infty,~if~C_i=ground~AND~H^b_j<H^b_i,\exists C_j\\
&k_{N+1},~if~C_i\neq ground~AND~S_i=N+1,\\
&0,~otherwise 
\end{aligned}
\right.
\label{eq:sp_cons}
\end{equation}
where $k_{N+1}$ is an integer which corresponds to the cost of an invisible support.

The physical constraint energy $E_{PC}$ is a weighted sum of $C_C$, $C_{dist}$ and $C_{SP}$ because they have different influences in practice. The formal expression is:
\begin{equation}
\begin{split}
E_{PC}(S,ST,C) = &\alpha_{C}C_C(S_i,C_i) + \alpha_{dist}C_{dist}(S_i,C_i,ST_i)\\
 &+ \alpha_{SPC}C_{SPC}(S_i,C_i).
\end{split}
\end{equation}
The optimal support relations are achieved by tuning the weights.

\begin{figure}[t!]
	\centering
	\framebox{
\includegraphics[width = 0.46 \textwidth]{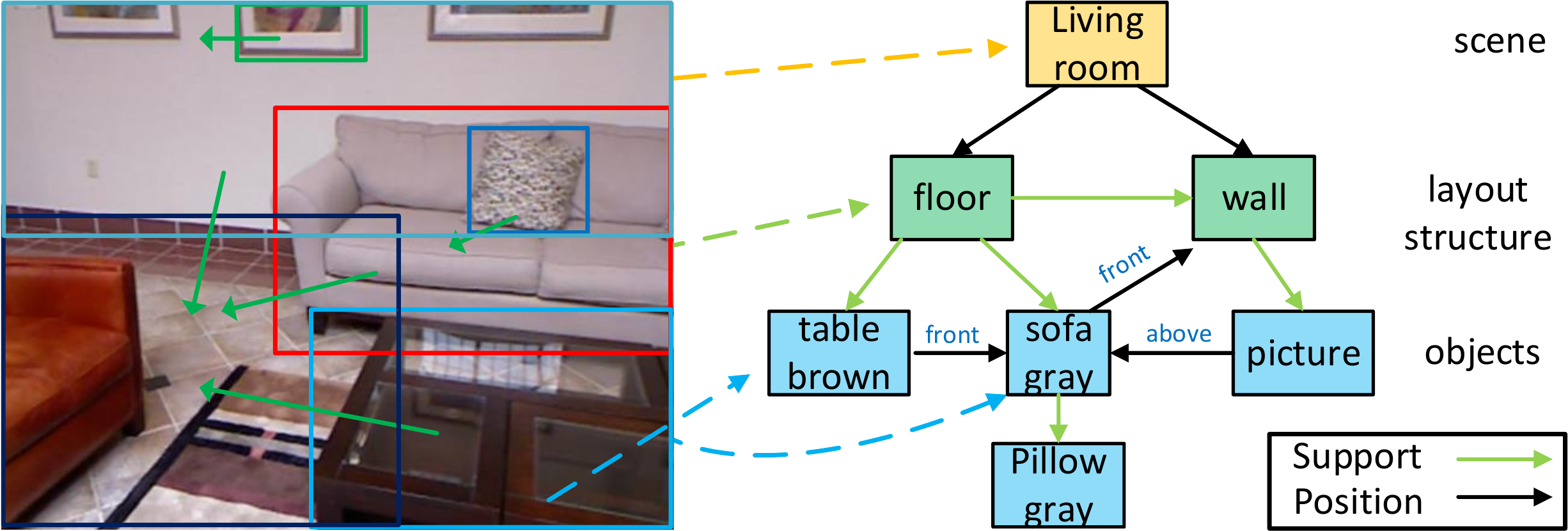}
}
\caption{An example of interpreting the image using our scene graph. In the root layer there is only one node to indicate the scene type; the first layer contains structure element ground, wall, ceiling and a hidden object; the lower layer contains other objects detected in the image except ground, wall, ceiling and hidden. Each node represents an individual object and each edge represents the support relation or relative position.}
\label{fig:sg}
\end{figure}

\subsection{Energy minimization}
The minimization of the energy function Eq. \eqref{eq:obj_func} can be formulated as an integer programming problem. Let $N^* =N+1$ indicate the number of detected objects plus the hidden supports. 
The Boolean indicator variable $B_{SP_{i,j}}$: $1\leq i\leq N, 1\leq j\leq 2N*+1$ encodes object $C_i$, its supporting objects $C_j$ and $i\neq j$ and support type $STi$. $B_{SP_{i,j}}=1, 1\leq j\leq N^*$ means that object $C_j$ supports object $C_j$ from behind. If $N^*+1\leq j\leq 2N^*$, then object $C_i$ is supported by $C_{j-N^*}$ from below, and $j=2N^*+1$ indicatges that $C_i$ is the ground and need no support. 
Boolean variable $B_{C_{i,\lambda}} = 1$ indicates that object $C_i$ has a class value $\lambda$.
Furthermore, variable $\chi^{\lambda,\upsilon}_{i,j}$ encodes the case $B_{SP_{i,j}}=1,B_{C_{i,\lambda}} = 1,B_{C_{j,\upsilon}} = 1$. The minimum energy inference problem is formulated as an integer program using this over-complete representation:
\begin{equation}
\argmin_{B_{SP},B_C,\chi} \sum_{i,j} \theta ^{SP}_{i,j}B_{SP_{i,j}} + \sum_{i,\lambda}\theta^C_{i,\lambda} B_{C_{i,\lambda}}+\sum_{i,j,u,v}\theta^\omega_{i,j,\lambda,\upsilon}\chi^{\lambda,\upsilon}_{i,j}.
\label{eq:integer_eq}
\end{equation}
In this formulation, the support energies $E_{SP}$ in Eq.~\eqref{eq:energy.SP} and the distance constraints $C_{dist}$ in Eq.~\eqref{eq:dist_cons} are encoded by $\theta ^{SP}_{i,j}$; the object class energies $E_C$ and support constraints $C_{SPC}$ in Eq.~\eqref{eq:sp_cons} are encoded by $\theta ^{C}_{i,\lambda}$; the object class constraints $C_C$ in Eq.~\eqref{eq:obj_class_cons} are encoded by $\chi^{\lambda,\upsilon}_{i,j}$.

The support constraints $C_{SPC}$ are enforced by
\begin{align}
\sum_jB_{SP_{i,j}}=1,~\sum_jB_{C_{i,\lambda}}=1,~\forall i,\\
\sum_{j,\lambda,\upsilon}\chi^{\lambda,\upsilon}_{i,j} = 1,~\forall i,\\
B_{SP_{i,j}}= B_{C_{i,\lambda}},~for~j = 2N^*+1,\lambda=1,~\forall i.
\end{align}

To ensure the definition of $\chi^{\lambda,\upsilon}_{i,j}$ and satisfy the object class constraints $C_C$, we require that
\begin{align}
\sum_{j,\lambda,\upsilon}\chi^{\lambda,\upsilon}_{i,j} = B_{SP_{i,j}},~\forall i,j\\
\sum_{j,\lambda,\upsilon}\chi^{\lambda,\upsilon}_{i,j} \leq B_{C_{i,\lambda}},~\forall i,\lambda.
\end{align}

The solution of the nteger program is defined as:
\begin{equation}
B_{SP_{i,j}},~B_{C_{i,\lambda}},~\chi^{\lambda,\upsilon}_{i,j}\in\{0,1\},~\forall i,j,\lambda,\upsilon.
\label{eq:np_hard}
\end{equation}
To solve Eq. \eqref{eq:np_hard} is an NP hard problem. Therefore, we relax this equation as:
\begin{equation}
B_{SP_{i,j}},~B_{C_{i,\lambda}},~\chi^{\lambda,\upsilon}_{i,j}\in[0,1],~\forall i,j,\lambda,\upsilon.
\label{eq:LP}
\end{equation}
Equation~\eqref{eq:LP} is a linear program and can be solved by the LP solver of the Gurobi package.

\begin{algorithm}[t]
\caption{Semantic Scene Graph Construction}
\label{Algo:Scene.Graph.Construction}
\begin{algorithmic}[1]
\STATE \textbf{Initialization}:
\STATE \textit{root} $\gets$ scene type
\STATE \textit{L} $\gets$ root
\WHILE{there are unassigned objects}
\WHILE{$L\neq 0$}
\STATE \textit{parent} $\gets$ first element of L
\STATE \textit{Remove} first element of L
\FOR{each object supported by \textit{parent}}
\STATE \textit{Create} node with 
\STATE \textit{Assign} to parent
\STATE \textit{Append} L $\gets$ object 
\ENDFOR
\ENDWHILE
\STATE \textit{Assign} renaming objects to hidden node
\ENDWHILE
\FOR{i=1:N}
\FOR{j=i:N}
\STATE \textit{Connect} $v_i$ and $v_j$ with edge $e_{i,j}$
\ENDFOR
\ENDFOR
\end{algorithmic}
\end{algorithm}

\begin{figure*}[t!]
	\centering
	\framebox{
	\includegraphics[width=0.7\textwidth]{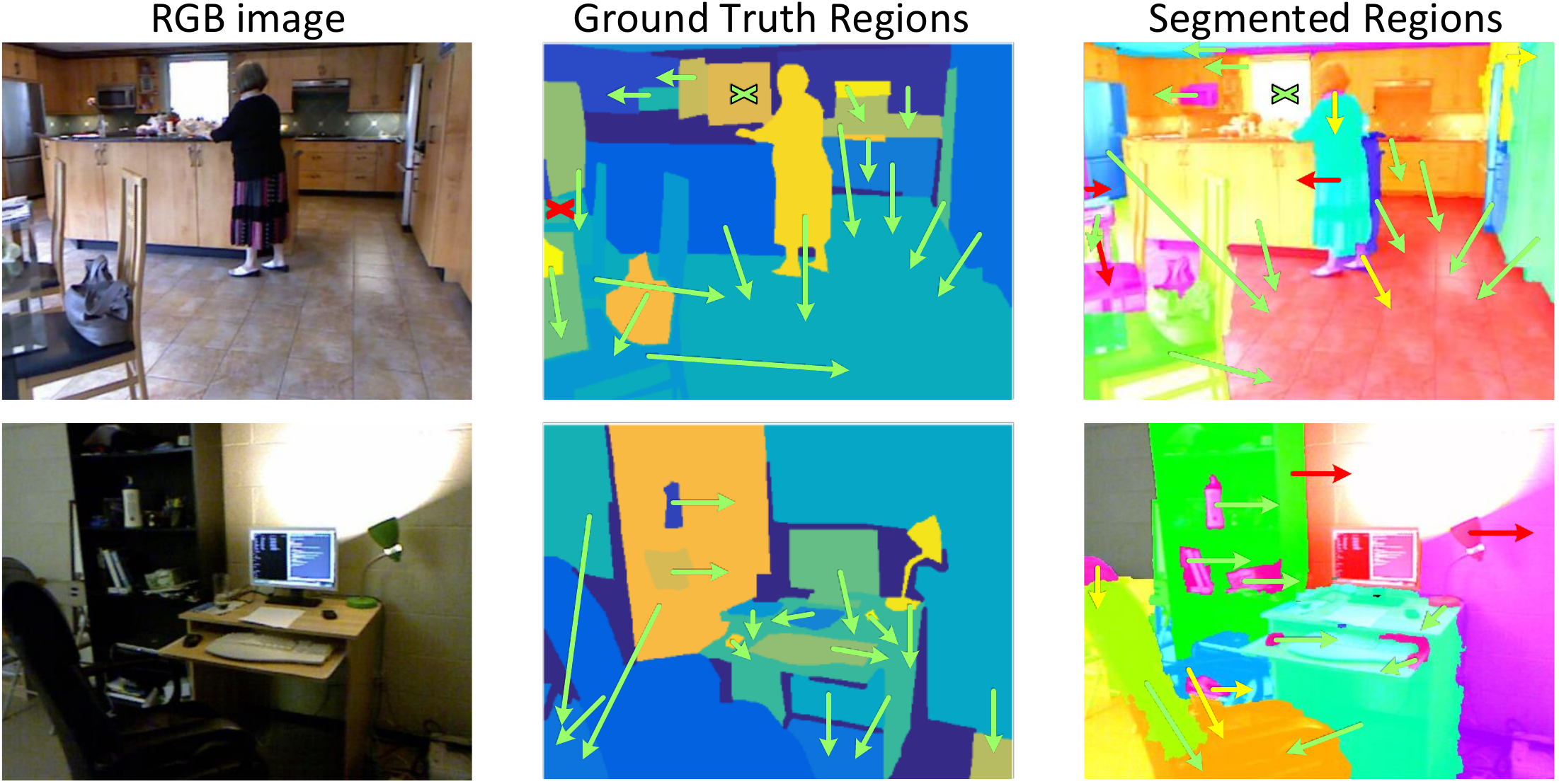}
	}
	\caption{Examples of support inference and object recognition. The middle column shows the results on the ground truth while the results in the last column are based on object detection. The direction of arrow indicates the supporting object. Cross denotes a hidden support. Correct support predictions in green and incorrect in red. The yellow ones mean that the support predictions are subjectively correct, but not exists in the ground truth. In the last column, objects belonging to the same class are denoted in the same color.}
	\label{fig:support_gt_seg}
\end{figure*}
\section{Scene Graph Construction}
\label{Sec:Scene.Graph}
Given a set of detected objects $C = \{C_1,\dots,C_N\}$, the object positions $P= \{p_1,\dots,p_N\}$, attributes $A= \{A_1,\dots,A_N\}$, and relationships between objects $R = \{R_{i,j}, i\neq j\}$, a scene graph $G$ is defined as a tuple $G = (V, E)$ where $V$ is the set of vertices and $E$ the set of edges, respectively. The triple $v_i \sim \{c_i,p_i,A_i\}$ represents object class $c_i$, position $p_i$ in the scene and attributes $A_i$ such as color, shape etc.
We train classifiers for the 8 most familiar colors in our live: red, green, blue, yellow, brown, black, white and gray using RGB features to recognize the object color. 
The position information is $p_i = (b_i,z_{min},z_{max})$, where $b_i$ defines the bounding box of the object and $(z_{min},z_{max})$ are its minimal and maximal depth respectively.
The relationships $\emph{R}_{i,j}$ represent support relations $T_{i,j}$ or relative positions between between objects.

Fig.~\ref{fig:sg} shows an example of a scene graph construction for a given image in the left. It is constructed using support relations as explained in Algorithm~\ref{Algo:Scene.Graph.Construction}. Please notice that a \emph{hidden} vertex is added to the second layer, i.e.~it is connected to the root node. Its purpose is is that unsupported objects can be assigned to it. Furthermore, walls are supported by the ground by default, and the ceiling by the walls. For indoor scenes, these are reasonable assumptions.

In the next step, we need to connect objects so as to describe the relative position of two nodes. 
For each object, we only define spatial information for objects which are close instead of creating a fully connected graph. 
For example in Fig.~\ref{fig:sg}, the brown table can be described as in front of the sofa or in front of the wall. 
But the former one is more exact to describe this spacial relation and more useful for scene understanding than the latter. In this paper, we describe the relative position with concepts above/under, front/behind and right/left, and each pair of them are symmetric. 


\section{Experiments and Results}
\label{Sec:Exps}
\subsection{Dataset}
Our experiments were conducted on the NYUv2~\cite{silberman2012indoor} dataset.
This dataset consists of 1449 images in 27 indoor scenes, and each sample consists of an RGB image and a depth map. 
The original dataset contains 894 object classes with different names. 
But some of them are similar classes (e.g. table and desk), or present rarely in the whole dataset, which is difficult to manipulate in practice. 
Therefore, we merge the similar object classes and discard those that appear sporadically. 
Finally, the object set consists of 32 classes and 3 generalized classes as "other prop", "other furniture", and "other structure". 
The 9 most frequent scene types are selected and the rest are generalized into the 10th type of "others".

The dataset is partitioned into two disjoint training and testing subsets, using the same split as \cite{silberman2012indoor}. 
For evaluation of the generated scene graphs, we manually built a scene graph for each scene based on ground truth using our GUI.


\subsection{Evaluating Support Relations}
Object segmentation is the foundation of support relation inference. Therefore, we evaluate the proposed method on both the ground truth segmentation and our segmentation which is based on object recognition. 
In the case of an object being detected without any other objects next to it, this object is assigned to be supported by the nearest surface, and its supporting object class is counted as hidden.
For some objects with complex shape and configuration, such as a corner cabinet being hung on the walls, the support prediction is counted as correct whichever wall is its support.
We also compare the experimental results with the best results of the most related work \cite{silberman2012indoor,zheng2015scene}.

\begin{table}[b]
    \caption{Results of the different approaches to predict support relations.The Accuracy is measured by total support relations which is correctly inferred divided by the number of objects in ground truth. The abbr. seg. is segmentation.}
    \label{tab:exp_comparison}
\centering
    \begin{tabular}{ c | c| c | c }
    \hline
    \multicolumn{4}{c}{Predicting Accuracy without Support Type}\\
    \hline
    Region Source 						 & ground truth &initial seg. & object seg.\\ 
    \hline
    Silberman \cite{silberman2012indoor} &	75.9 	& 54.1	& 55.1 \\
    \hline
    Xue \cite{zheng2015scene} 			 &	77.4	& 56.2	& 58.6 \\
    \hline
    Ours					  			 & 88.4     &$\setminus$ & 65.7 \\
    \hline
    \multicolumn{3}{c}{Predicting Accuracy with Support Type}\\
    \hline
    Silberman \cite{silberman2012indoor} & 72.6 	& 53.5	& 54.9 \\
    \hline
    Xue \cite{zheng2015scene} 			 & 74.5		& 55.3	& 56.0 \\
    \hline
    Ours					  			 & 82.1     &$\setminus$	& 61.5 \\
    \hline
    \end{tabular}
    \vspace{0.3cm}
\end{table}
The experimental results and the comparison are listed in Tab.~\ref{tab:exp_comparison}.
The  accuracy of support relations predictions differentiate between without and with support type. When the support type is not considered, the predicted support relations which have correct supporting and supported objects are counted as correct. When the support type is taken into account, only when the predicted support type (from behind or below) is also correct, this prediction is accounted to be correct. 
When using the ground truth, our method of using contextual relations between object categories outperforms using only 4 structure categories. 
It demonstrates the effectiveness of contextual relation between different classes of objects to understand their support relations. 
Comparing the results based on the ground truth and object recognition given by our approach, the latter performance drops about $23\%$. It explains that the accuracy of object recognition is the main limitation for understanding support relations between objects. 
This conclusion is verified again by the results given by~\cite{silberman2012indoor} and \cite{zheng2015scene} comparing the results based on the ground truth and their segmentation.
 
Our approach does not achieve significant boosting in support relation prediction using object segmentation against the other two works (the last column in Tab.~\ref{tab:exp_comparison}) for two reasons: 1) they used only 4 structure classes while our works detect 32 concrete object categories. We do not achieve very good results in object recognition; 
2) the pixel-wise segmentation is more advantageous in estimating the spatial information about objects compared with our coarse segmentation. 
Furthermore, pixel-wise segmentation ensures that each object in given image has at least one support relation with one of its neighbor objects, while in our approach some objects are not detected, especially in a cluttered scene. 

Nevertheless, only using the 2D bounding box and superpixel maps is faster than previous  works. 
In contrast, our method segments objects using the ready-made bounding boxes and superpixel maps provided by RCNN. 
No mutually call by each other of support relation and image segmentation \cite{silberman2012indoor,xue2015towards}, which is the one of the main novelty in the other two works. 
Comparing the results of the two works between initial segmentation (the 3rd column) and improve object segmentation (the last column), their improved segmentation do not improve the predicting accuracy of support relations too much. 

\begin{figure}[t]
	\framebox{
	\centering
	\subfigure[Dense object detection]{
	\label{subfig:dense_bb}
	\includegraphics[width=0.225\textwidth]{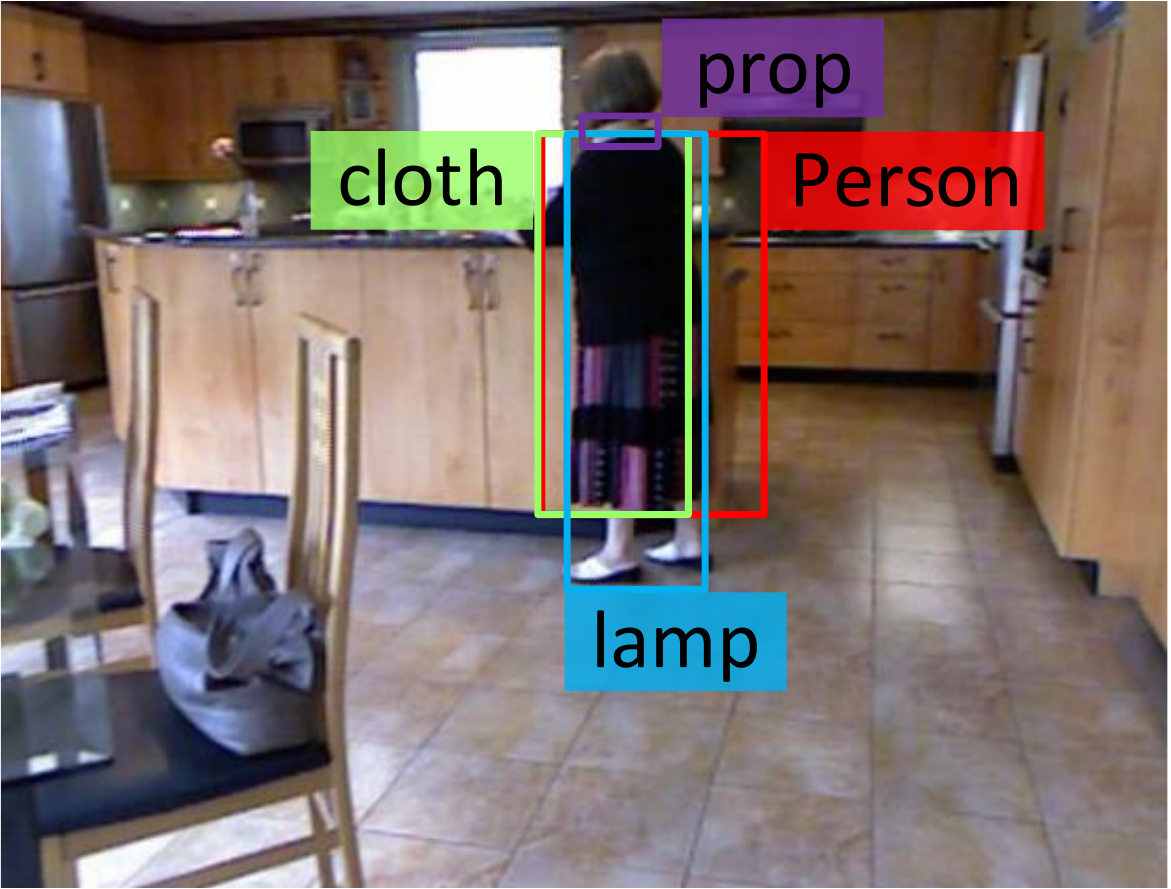}}
	\subfigure[Confused detection]{
	\label{subfig:bed_counter}
	\includegraphics[width=0.225\textwidth]{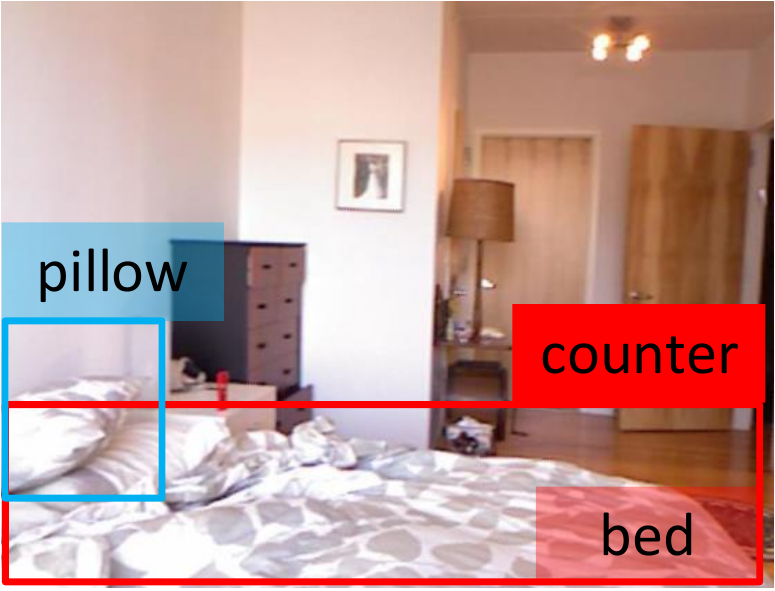}}
	}
	\caption{Examples of imperfect object detection based on bounding box.}
\label{fig:bed_bb}
\end{figure}
Visual examples are shown in Fig.~\ref{fig:support_gt_seg}. From the middle column we can see that, our approach performs well on the ground truth. The last column also illustrates the object recognition and segmentation. 
Some objects are not detected in the second row: the screen, keyboard and other props. 
Their regions are merged into other objects, because the bounding boxes involve them, e.g. the screen belongs to the wall region and the keyboard belongs to the table surface. 
The lamp is falsely considered to be supported by the wall because its joint lever is not detected. 
Another problem is that a complete object is sometimes recognized as multiple objects.
For instance,
the woman in the upper row is divided into 4 parts, cloth (cyan), body (blue), the neck is detected as a prop (pink) and the feet are detected as part of a lamp (light pink). It leads to incorrect support inference of the cloth being supported by the nearest cabinet. This phenomena is caused by dense detection in the regions of the person, as shown in Fig.~\ref{subfig:dense_bb}. 

Because our inference is jointly minimized by the energy function Eq.~\eqref{eq:energy}, the final results of the object recognition accuracy are improved. For instance as shown in~Fig.~\ref{subfig:bed_counter}, the white bed with large flat surface is recognized as counter and bed in the same bounding box. 
Due to a pillow supported by this confused object, it leads the inference to decide it as bed, because a pillow is rarely supported by a counter in the prior knowledge.

\subsection{Evaluating Scene Graph}
\begin{figure}[t]
	\centering
	\framebox{
\includegraphics[width=.2\textwidth]{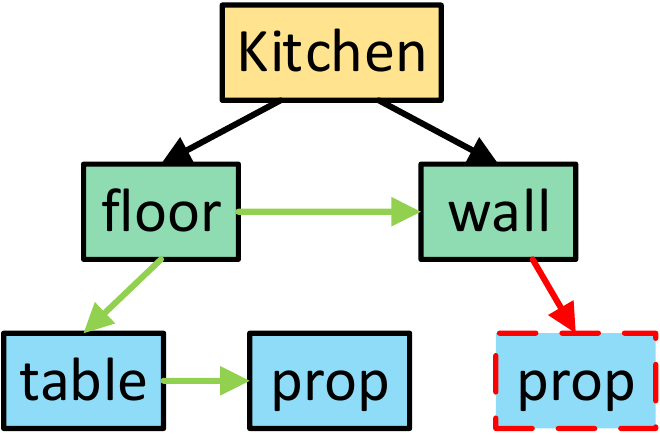}
}
\caption{An example of scene graph with an error (in red).}
\label{fig:sg_error}
\end{figure}
\begin{table}[t]
\caption{Matrix of the scene graph. The red numbers indicate the errors from the above graph. }
\label{tab:sg_matrxi}
\centering
\begin{tabular}{l r| c c c c c |}
 & \multicolumn{1}{r}{} & \multicolumn{5}{c}{Supporting}\\
\parbox[t]{2mm}{\multirow{7}{*}{\rotatebox[origin = c]{90}{Supported}}} & \multicolumn{1}{c}{} & \multicolumn{1}{c}{kitchen} & \multicolumn{1}{c}{floor} & \multicolumn{1}{c}{wall} & \multicolumn{1}{c}{table} & \multicolumn{1}{c}{prop}\\
\cline{3-7}
&kitchen &0		  &	1		&	1	&	0	&	0		\\
&floor	&1		  &	0		&	0	&	0	&	0		\\
&wall	&1		  &	0		&	0	&	0	&	0		\\
&table	&0		  &	1		&	0	&	0	&	0		\\
&prop	&0		  &	0		&	0/{\color{red}{1}}	&	1/{\color{red}{0}}	&	0		\\
\cline{3-7}
\end{tabular}
\end{table}
Because our goal is to evaluate the structure quality of generated scene graph, the attributes of objects and relative positions between objects are not taken into account in this work. We represent the directed, unweighted graph by its affinity matrix, as shown in Fig.~\ref{fig:sg_error} and Tab.~\ref{tab:sg_matrxi}. Here, the object class corresponding to the columns supports the classes corresponding to the rows. 

To measure the similarity with ground truth graphs, we create graphs  $G^\prime ( V^\prime, E^\prime )$ with undirected edges such that $V^\prime = V$ and $E^\prime = \{ e^\prime_{ij}=1 \; \Leftrightarrow \; e_{ij} = 1 \; \vee \; e_{ji} = 1 \}$. 
We do so for both the estimated scene graph and the ground truth. To estimate their similarity, we compare their Cheeger constants. The Cheeger constant $h_G$ of a graph is defined to be $h_G = \min_S h_g(S)$~\cite{Chung1996:Spectral}. Here, $S$ denotes a subset of the vertices of $G$, and 
\begin{equation}
h_G(S) = \frac{ \left| E (S,\bar{S}) \right|}{\min \left(\mbox{vol} S, \mbox{vol} \bar{S} \right)} 
\end{equation}
with $\mbox{vol} S = \sum_{x \in S} d_x$ being the volume of $S$. $d_x$ is the degree of vertex $x$, and $\bar{S}$ is the complement set of $S$, i.e. $\bar{S} = V \setminus S$. The symbol $| \cdot |$ indicates the cardinality. Since $h_G$ is hard to compute, we use upper and lower bounds
\begin{equation}
l_G = \frac{1}{2}(1-\lambda_2) \leq h_G \leq \sqrt{2-2\lambda_2} = u_G.
\label{Eq:Similarity.Cheeger}
\end{equation}
Here, $\lambda_2$ denotes the second largest eigenvalue of the random walk matrix $P(G) = D^{-1} A$ of the graph $G$ with affinity matrix $A$ and $D_{ii} = d_x$. We then take $| (u_G^\prime-l_G^\prime)-(u_{H^\prime}-l_{H^\prime})|$ as similarity between $G^\prime$ and the undirected graph $H^\prime$ corresponding to the ground truth graph $H$.

Since there may be incorrectly estimated scene graphs whose Cheeger constants nonetheless do not differ from those of the ground truth graph\footnote{Please refer to the supplementary material for an example.}, we take
\begin{equation}
    \left\| u_2(G^\prime) \cdot u_2(G^\prime)^\top - u_2(H^\prime) \cdot u_2(H^\prime)^\top \right\|_F / \sqrt{|V(H^\prime)|} 
    \label{Eq:Similarity.Cut}
\end{equation}
as further measure of the similarities between the two graphs. In Eq.~\eqref{Eq:Similarity.Cut}, $u_2$ denotes the eigenvector corresponding to $\lambda_2$ and $\| \cdot \|_F$ the Frobenius-norm, $| \cdot |$ the cardinality, and $\sqrt{|V(H^\prime)|}$ is for normalization.

Lastly, we use a naive heuristic to measure the difference between the two graphs. Its computation is explained in the supplementary material. 

\begin{table}[t!]
    \caption{Results of different measures to evaluate the quality of generated scene graph comparing with ground truth. The number is smaller, the quality is better. 0 means the generated graph is identical with the ground truth.}
    \label{tab:exp_support}
\centering   
    \begin{tabular}{| c | c | c | c |}
    \hline
    \multicolumn{4}{|c|}{Evaluation of generated scene graph}\\
    \hline
    Measures 		& Cheeger \eqref{Eq:Similarity.Cheeger} & Spectral \eqref{Eq:Similarity.Cut} & Naive \\ 
    \hline
    \hline
    Mean 	 		& 0.19  					 & 0.20 				 & 0.41 \\
    \hline
    Variance		& 0.16 						 & 0.07 			     & 0.15  \\
    \hline
    \end{tabular}
\end{table}

The evaluation results using the three different measures on the test dataset are shown in Tab.~\ref{tab:exp_support}. We can see that the generated scene graphs achieve low mean error values when comparing with ground truth. It proves that our method generates reasonable scene graph given a scene.
\section{Conclusion}
\label{Sec:Conclusion}
This work presents a new approach for inferring accurate support relations between objects from given RGBD images of cluttered indoor scenes. 
We also introduce how to construct semantic scene graphs interpreting physical and contextual relations between objects and environment. 
This topic is a necessary step for deeper scene understanding.
%
The proposed framework takes RGBD images as input, detects object using RCNN and then conducts a simple object extraction from a superpixel map, which is faster then pixelwise segmentation.
Next, reasonable support relations are inferred by using physical constraints and prior support knowledge between object classes.
Finally, support relations, along with contextual semantics, scene recognition, and object recognition allow to infer a semantic scene graph.
Using the NYUv2 dataset, the inferred support relationships are more accurate than those achieved from previous algorithms. 
For assessing the semantic scene graphs, ground truth graphs are created, and objective measures for graph comparison are proposed. Evaluation results show that the inferred scene graphs are reasonable.
The ground truth graphs and the tool to create them will be public available.

In future research, we will experiment on letting robot finish specific tasks using our scene graphs. Furthermore, it would be nice to give the support relations between objects semantic meaning, e.g. "the man is sitting in the sofa". 
From the scene graph, it would be interesting to group objects into meaningful groups, such as studying area. The scene graphs also should improve the prediction accuracy of support relations and object classes in an iterative manner. 
At last, we will improve the measures to assess the quality of the scene graph hypotheses.


\bibliographystyle{IEEEtran}
\bibliography{egbib}

\section{Supplementary Material}

\subsection{Support Features}
\begin{figure*}
\centering
\includegraphics[width=0.8\textwidth]{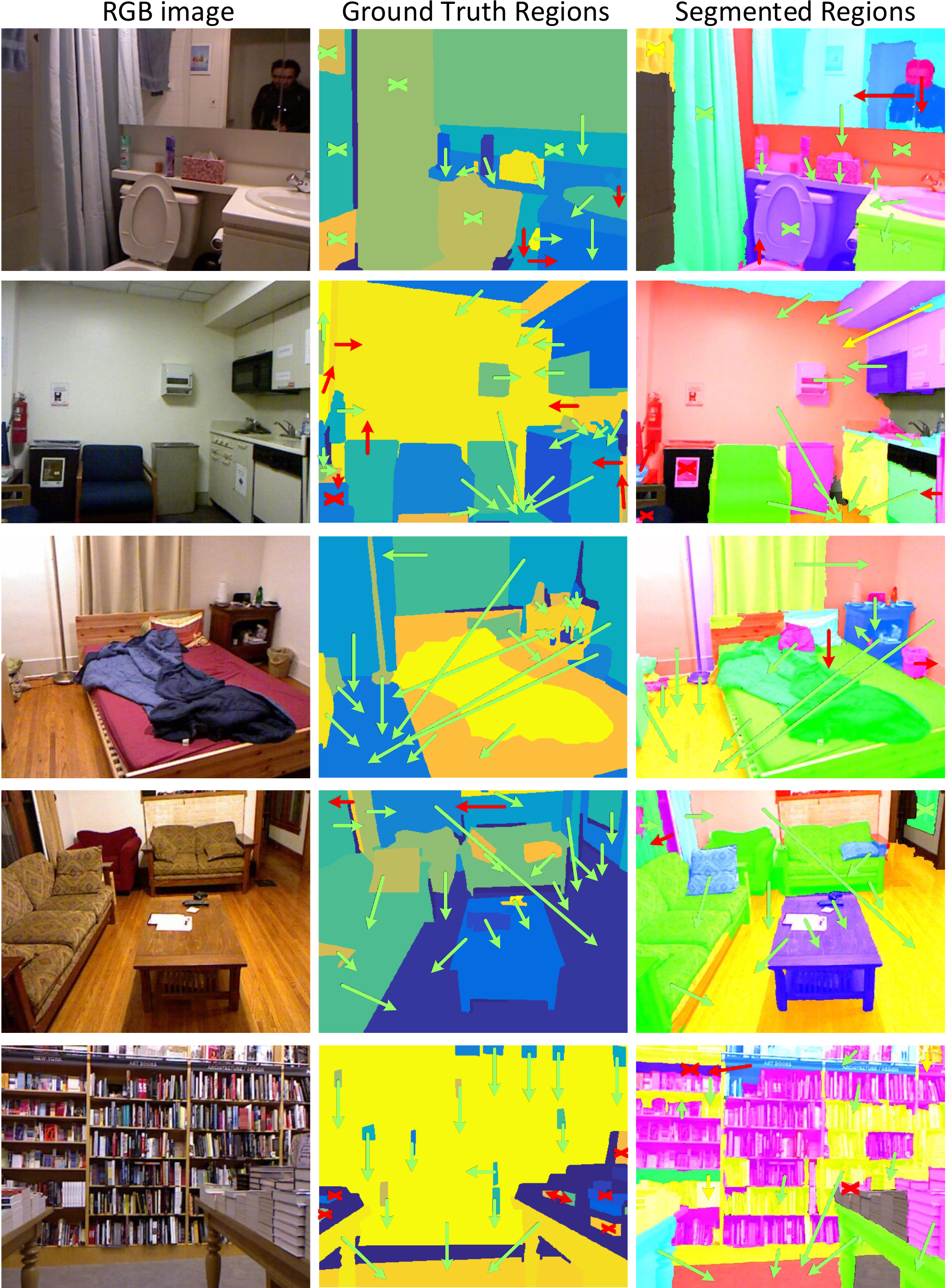}
	\caption{Examples of support and object class inference with the LP solution. The middle column shows the results on the ground truth while the results in the last column are based on object detection. The direction of arrow indicates the supporting object. Cross denotes a hidden support. Correct support predictions in green and incorrect in red. The yellow ones mean that the support predictions are subjectively correct, but the object detection or segmentation are incorrect. In the last column, objects belonging to the same class are denoted in the same color.}
	\label{Fig:support}
\end{figure*}

Support features are listed in Table~\ref{tab:support_features}.
\begin{table*}
\centering
\begin{tabular}{|p{11.5cm}|l|}
\hline
Support Feature Description & Dims\\
\hline
\hline
\textbf{Geometry} & \textbf{8} \\
G1. Minimum vertical and horizontal distance between the two regions & 2\\
G2. Absolute distance between the regions' centroids & 1\\
G3. The lowest heights of the two regions above the ground & 2\\
G4. Percentage of the supporting region that is farther from the viewer than the supported region & 1 \\
G5. Percentage of the supported region contained inside convex hull of supporting region's projection onto the floor plane & 1\\
G6. Percentage of the supported region contained inside convex hull of supporting region's horizontal points when projected onto the floor plane & 1\\
\hline
\textbf{Shape} & \textbf{9}\\
S1. Number and percentage of horizontal pixels in the supporting region & 2 \\
S2. Number and percentage of horizontal pixels in the supported region & 2 \\
S3. Number and percentage of vertical pixels in the supporting region & 2 \\
S4. Number and percentage of vertical pixels in the supported region & 2 \\
S5. Chi-squared points when projected onto the floor plane			& 1 \\
\hline
\textbf{Region} &\textbf{3}\\
R1. Ratio of number of pixels between the supported supported and supporting region & 1\\
R2. Whether or not the two regions are neighbors in the image plane & 1\\
R3. Whether or not the supporting region is hidden & 1\\
\hline
\end{tabular}
\caption{Support feature description.}
    \label{tab:support_features}
\end{table*}
Some examples of inferring support relations in given images are shown in Fig.~\ref{Fig:support}.

\begin{table}[b!]
\centering
    \begin{tabular}{| l | l |}
    \hline
    Relative Position & ~~~~~~~~~~~~~~~Condition \\ \hline
    Above & $y^{min}_i>y^{max}_j$; $z^{min}_j<z^{min}_i<z^{max}_j$; $I^x_{i,j},I^z_{i,j}\neq \emptyset$ \\ \hline
    Behind 1& $z^{min}_i\geq z^{max}_j$; $I^x_{i,j}\neq \emptyset$ \\ \hline
    Behind 2& $\frac{1}{2}(z^{min}_i+z^{max}_i)>z^j_{max}$; $I^x_{i,j}\neq \emptyset$ \\ \hline
    Right & $z^j_{min}<\frac{1}{2}(z^{min}_i+z^{max}_i)<z^j_{max}; \frac{1}{2}(x^{min}_i+x^{max}_i)>x^j_{max}$ \\
    \hline
    \end{tabular}
    \caption{The decision rules of object position relative to an object.}
    \label{tab:relative_position}
\end{table}

\subsection{Relative Position Decision}
\begin{figure*}[htp]
	\centering
	\subfigure[above]{
	\label{subfig:above}
	\includegraphics[width=0.18\textwidth]{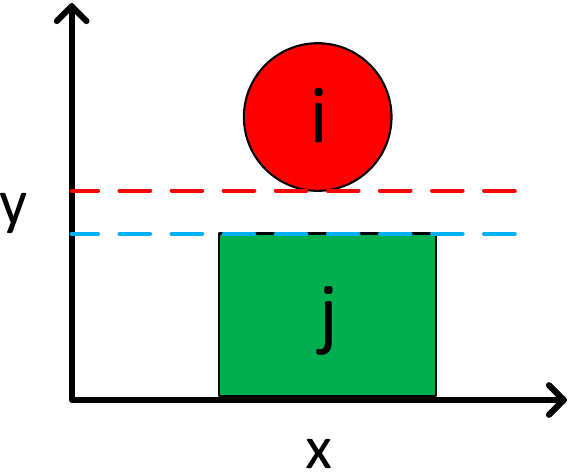}
	\includegraphics[width=0.18\textwidth]{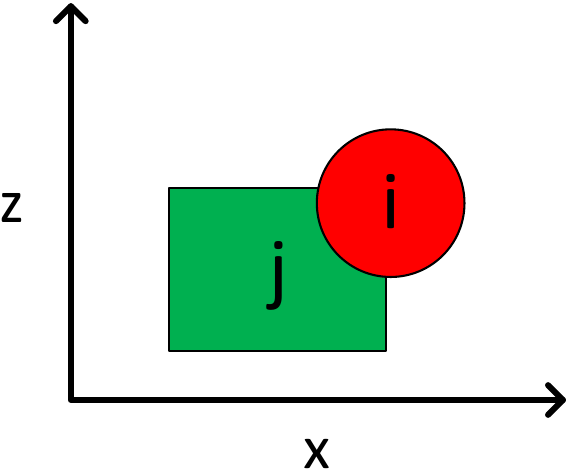}}
	\subfigure[behind 1]{
	\label{subfig:behind_1}
	\includegraphics[width=0.18\textwidth]{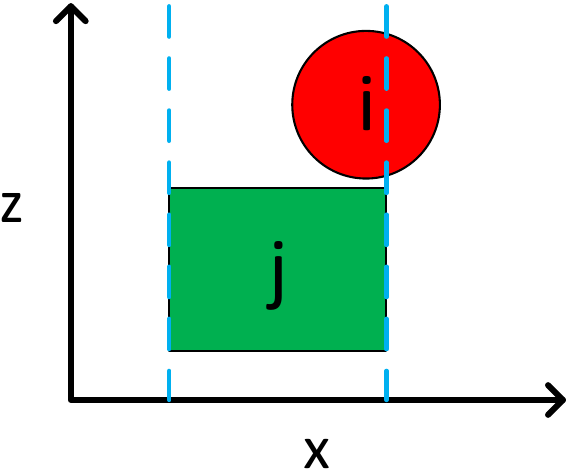}}
	\subfigure[behind 2]{
	\label{subfig:behind_2}
	\includegraphics[width=0.18\textwidth]{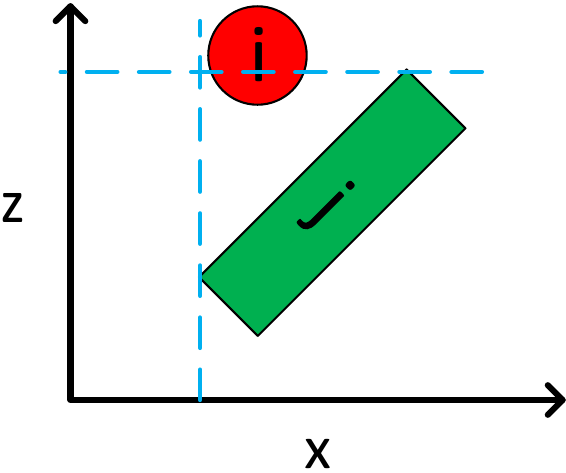}}
	\subfigure[right]{
	\label{subfig:right}
	\includegraphics[width=0.18\textwidth]{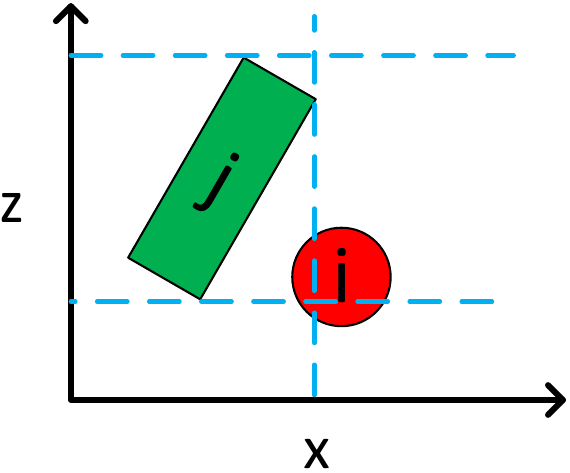}}
	\caption{Relative position described by (a) above; two case of behind in (b) and (c); and right in (d).}
\label{fig:relative_position}
\end{figure*}

For object $i$, its spatial information can be represented by 3D room coordinates $(x^i_{min},x^i_{max},y^i_{min},y^i_{max},z^i_{min},z^i_{max})$, where $x-z$ is the floor plane and $y$ direction points upward. 
For convenient discussion, we define some symbols here. $I^x_{i,j} =[x^{min}_i,x^{max}_i]\cap [x^{min}_j,x^{max}_j]$ denotes the intersection between object $i$ and $j$ when they are projected on the $x$ axis, and so do $I^y_{i,j}$ and $I^z_{i,j}$ too.
The rule for deciding object $i$ position relative to object $j$ is listed in Table~\ref{tab:relative_position} and illustrated in Fig.~\ref{fig:relative_position}. 

Because the position of above-under, behind-front and right-left are symmetric, we don't describe the versa here any more.

\begin{figure*}
\centering
\includegraphics[width=0.85\textwidth]{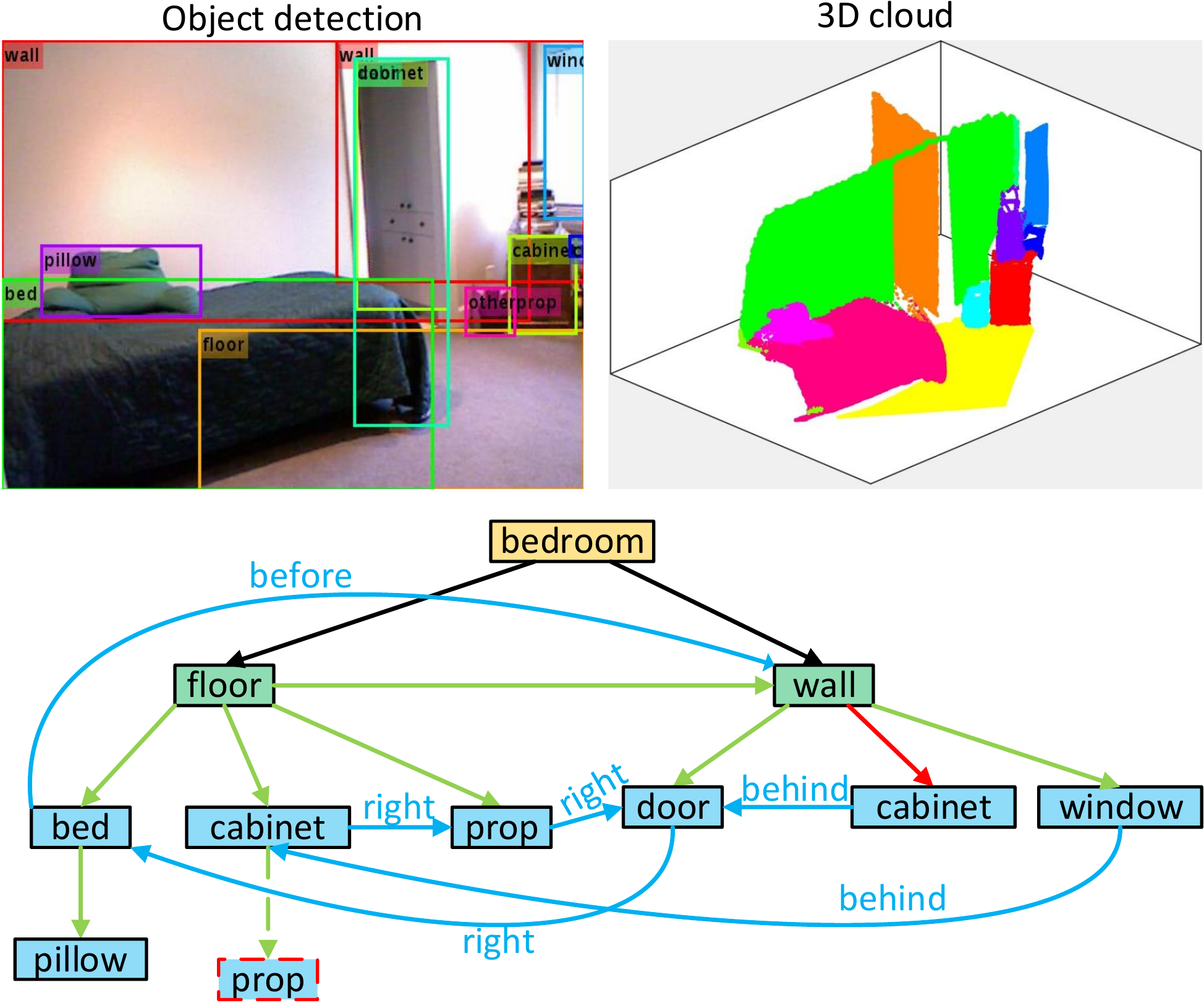}
\caption{An example of constructed scene graph based on object classes, support relations inference and relative position estimation. The up-right image shows the results of object detection and the up-left image shows the 3D points cloud estimated from the RGBD image. In the graph, the green arrows point to the supported object and the blue  paths with arrows indicate the relative position. The red arrow indicates the incorrect inferred support relation. The missed detected object is shown in a box with red dashed frame and its support relation is expressed by a dashed green arrow.}
\end{figure*}

\subsection{Evaluating Scene Graph}

\subsubsection{Laplacian Measure}
\begin{figure*}[ht!]
	\centering
	\subfigure[Ground truth scene graph]{
	\label{subfig:gt_sg}
	\includegraphics[width=0.34\textwidth]{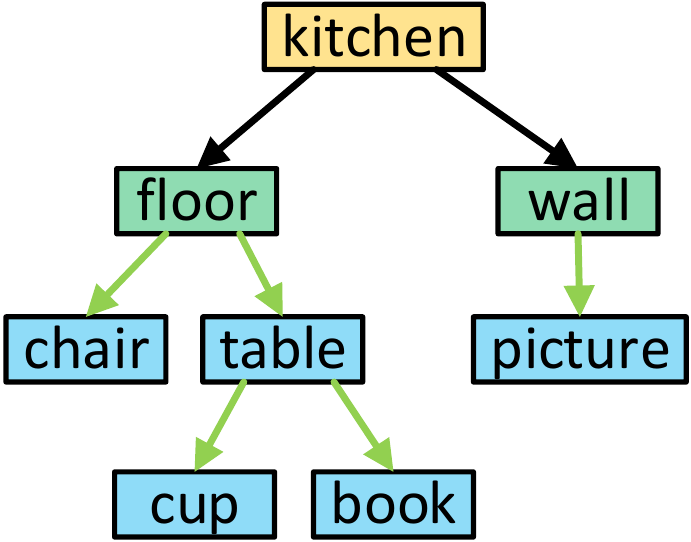}}
	\hspace{0.1cm}
	\subfigure[Scene graph with error]{
	\label{subfig:er_sg}
	\includegraphics[width=0.4\textwidth]{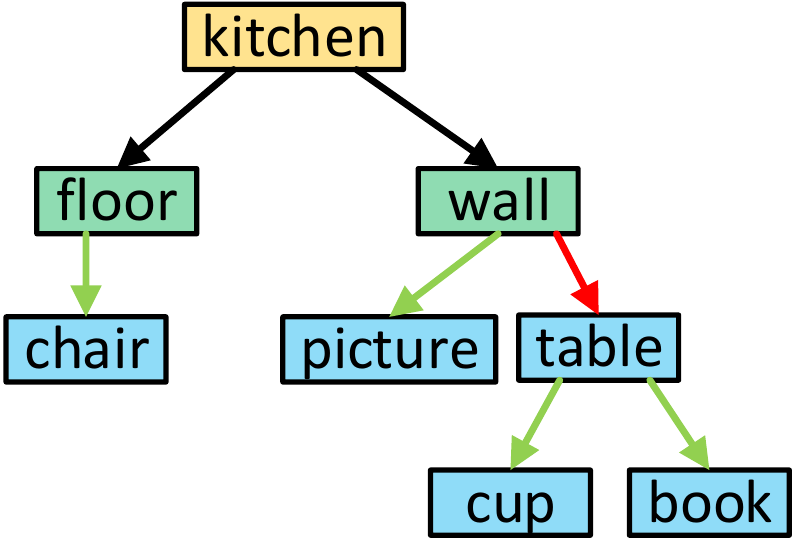}}
	\caption{An examples of constructed scene graph (b) comparing with its ground truth (a). The incorrect support relation is indicated by a red arrow.}
\label{fig:bed_bb}
\end{figure*}
Consider the examples of scene graphs shown in Fig.~\ref{fig:bed_bb}. In the left plot, a ground truth scene graph is shown. The right image in the same figure shows an example of an estimated graph. It can be easily seen that the subgraph with root being the \emph{table} vertex is incorrectly assigned to the \emph{wall}. In the following, consider the graphs relaxed so as to have undirected edges.

A graph measures based on an isoperimetry such as Eq.~(18) cannot capture this difference since the ratio between surface area and volume remains unchanged.

Therefore we further use a measure inspired by a normalized cut of each graph given by the eigenvector $u_2$ to the second smallest eigenvalue of the graph Laplacian. For the left graph the decision boundary induced by the hyperplane with normal $u_2$ cuts the graph between \emph{kitchen} and \emph{floor}, whereas it cuts between \emph{kitchen} and \emph{wall} for the graph shown in the right image of Fig.~\ref{fig:bed_bb}. A measure of the difference between the two hyperplanes is given by 
\begin{equation}
    \left\| P_{u_2(G_1)} - P_{u_2(G_2)} \right\|_F
    \label{Eq:Proj}
\end{equation}
where $u_2(G_i)$ denotes the second smallest eigenvector of the Laplacian of graph $G_i$, and $P_{u_2}$ the orthogonal projection onto $\mbox{span}\left( u_2(G_i) \right)$. Since the number of vertices can differ between images, we normalize Eq.~\eqref{Eq:Proj} by the maximum number of vertices the ground truth scene graphs can have.
\begin{table}
\centering
\begin{tabular}{l l r| c c c c c c c c|}
 & \multicolumn{2}{r}{} & \multicolumn{8}{c}{Supporting}\\
\multicolumn{2}{l}{\parbox[t]{2mm}{\multirow{7}{*}{\rotatebox[origin = c]{90}{Supported}}}} & \multicolumn{1}{c}{} & \multicolumn{1}{c}{kitchen} & \multicolumn{1}{c}{floor} & \multicolumn{1}{c}{wall} & \multicolumn{1}{c}{table} & \multicolumn{1}{c}{chair} & \multicolumn{1}{c}{picture} & \multicolumn{1}{c}{cup} & \multicolumn{1}{c}{book}\\
\cline{4-11}
&&kitchen &0	&1	&1	&0	&	0	&0	&0	&0	\\
&&floor	&1	&0	&0	&0	&	0	&0	&0	&0	\\
&&wall	&1	&0	&0	&0	&	0	&0	&0	&0	\\
&&table	&0	&1	&0	&0	&	0	&0	&0	&0	\\
&&chair	&0	&1	&0	&0	&	0	&0	&0	&0	\\
&&picture&0	&0	&1	&0	&	0	&0	&0	&0	\\
&&cup	&0	&0	&0	&1	&	0	&0	&0	&0	\\
&&book	&0	&0	&0	&1	&	0	&0	&0	&0	\\
\cline{4-11}
\end{tabular}
\caption{Matrix of the ground truth scene graph Fig.~\ref{subfig:gt_sg}.}
\label{tab:sg__gt_matrxi}

\centering
\begin{tabular}{l l r| c c c c c c c c|}
 & \multicolumn{2}{r}{} & \multicolumn{8}{c}{Supporting}\\
\multicolumn{2}{l}{\parbox[t]{2mm}{\multirow{7}{*}{\rotatebox[origin = c]{90}{Supported}}}} & \multicolumn{1}{c}{} & \multicolumn{1}{c}{kitchen} & \multicolumn{1}{c}{floor} & \multicolumn{1}{c}{wall} & \multicolumn{1}{c}{table} & \multicolumn{1}{c}{chair} & \multicolumn{1}{c}{picture} & \multicolumn{1}{c}{cup} & \multicolumn{1}{c}{book}\\
\cline{4-11}
&&kitchen &0&1	&1	&0	&	0	&0	&0	&0	\\
&&floor	&1	&0	&0	&0	&	0	&0	&0	&0	\\
&&wall	&1	&0	&0	&0	&	0	&0	&0	&0	\\
&&table	&0	&{\color{red}{0}}	&{\color{red}{1}}	&0	&	0	&0	&0	&0	\\
&&chair	&0	&1	&0	&0	&	0	&0	&0	&0	\\
&&picture&0	&0	&1	&0	&	0	&0	&0	&0	\\
&&cup	&0	&0	&0	&1	&	0	&0	&0	&0	\\
&&book	&0	&0	&0	&1	&	0	&0	&0	&0	\\
\cline{4-11}
\end{tabular}
\caption{Matrix of the constructed scene graph Fig.~\ref{subfig:er_sg}. The red numbers indicate the errors in the graph: the table is supported by wall instead of by floor.}
\label{tab:sg__err_matrxi}
\end{table}

\subsubsection{Heuristic}
Beside the two evaluation methods "Cheeger" Eq. (19) and "Spectral" Eq. (20) as described in the paper, we propose another naive method to measure the constructed graph. The matrices for describing the ground truth scene graph Fig.~\ref{subfig:gt_sg} and the constructed graph Fig.~\ref{subfig:er_sg} are shwon in Tab.~\ref{tab:sg__gt_matrxi} and Tab.~\ref{tab:sg__err_matrxi} respectively. The matrix can not only describe the support relations between object but also the object classification. The difference between matrices $M_i$ and $M_j$ is formally calculated as: 
\begin{equation}
d_{i,j} =  \frac{|M_i\oplus M_j|}{|M_i\lor M_j|}
\end{equation}
where $|.|$ is the total number of 1 in a matrix. Even though it is not a sophisticated method, it is a complementary measure for reference.

\subsection{GUI}
We provide a convenient GUI tool for generating ground truth graphs. Upon acceptance, the tool and the ground truth scene graph for NYUv2 dataset will be available on the authors' home page. Please turn to the video supplementary material to have a look at our GUI tool.

\end{document}